\documentclass[letterpaper, 10 pt, conference]{ieeeconf}  

\IEEEoverridecommandlockouts                              

\usepackage{graphics} 
\usepackage{epsfig} 
\usepackage{mathptmx} 
\usepackage{times} 
\usepackage{amsmath} 
\usepackage{amssymb}  
\usepackage{booktabs}
\usepackage{capt-of}

\usepackage{xspace}
\def\OURS{EgoCPT\xspace}

\newcount\Comments 
\usepackage{array,multirow,graphicx}
\usepackage{color}
\usepackage{makecell}
\usepackage[T1]{fontenc}
\usepackage{wrapfig}
\usepackage[hidelinks]{hyperref}
\usepackage[all]{hypcap}
 \usepackage{amsmath,amssymb}
\definecolor{darkgreen}{rgb}{0,0.5,0}
\definecolor{purple}{rgb}{1,0,1}
\newcommand{\figref}[1]{\hyperref[#1]{Fig.~\ref*{#1}}}
\newcommand{\kibitz}[2]{\ifnum\Comments=1\textcolor{#1}{#2}\fi}

\title{\LARGE \bf
\OURS: Learning Dexterous Manipulation from a Single Ego-Centric Video with  Contact-Aware Optimization
}

\author{Zhixin Sun$^{1}$, Kairui Shi$^{1, 2}$, Tianao Ren$^{3}$,\\ Ming Yang$^{1}$, Mark Cutkosky$^{3}$, James Jingxi Xu$^{1, 3}$
\thanks{$^{1}$Ant Group}%
\thanks{$^{2}$Computer Science and Engineering, University at Buffalo}%
\thanks{$^{3}$Mechanical Engineering, Stanford University}%
\\
\url{https://ego-cpt.github.io/}
}

\begin{document}

\maketitle
\thispagestyle{empty}
\pagestyle{empty}

\begin{abstract}

While expert demonstrations on real robots are difficult to collect, human ego-centric videos are easy to collect and abundant on the internet. However, learning dexterous manipulation policies from human videos remains challenging---one of the most important challenges being the difficulty of reconstructing physically grounded robot trajectories due to the large embodiment gap. In this paper, we propose \OURS, a novel paradigm for learning high-quality retargeted dexterous hand trajectories directly from an ego-centric video, which can then be deployed directly on real robots. The only inputs to our method are an ego-centric RGB video and a reference hand pose trajectory, which can be easily obtained through off-the-shelf software or specialized cameras during recording. Our method is particularly robust to noisy hand poses attributed to our novel contact-aware optimization framework, which iteratively optimizes the contact points between the robot hands and objects. Across five manipulation tasks, our method achieves overall success rates of 67.3\% on noisy EgoDex hand-pose reconstructions and 92.0\% on high-quality TACO reconstructions collected using motion-capture hardware, outperforming multiple baselines on both benchmarks.



\end{abstract}

\suppressfloats[t]
\begin{figure}[t]
  \centering
  \includegraphics[width=\columnwidth]{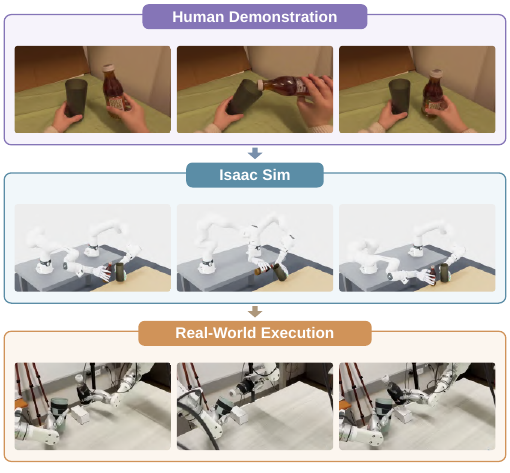}
  \caption{\textbf{From human demonstration to robot execution.} \OURS reconstructs robot-feasible hand--object interactions from an egocentric pouring demonstration. Top to bottom: human demonstration, policy execution in Isaac Sim, and deployment on the real robot.}
  \label{fig:teaser}
\end{figure}


\section{Introduction}
\label{sec:introduction}

Dexterous robot hands can produce rich contact behaviors, but learning such behaviors remains constrained by the cost of collecting robot-specific demonstrations. Teleoperation, motion capture, and instrumented gloves require specialized hardware, careful calibration, and substantial task-specific effort, while the resulting data are often tied to a particular robot embodiment. Human hand--object interaction data offer a complementary source of supervision. Existing datasets provide increasingly accurate hand poses, object geometry, object motion, and contact observations \cite{kwon2021h2o,fan2023arctic,banerjee2025hot3d,chao2021dexycb,wang2025hocap}, and modern egocentric devices can record synchronized RGB video and metric hand skeletons without instrumenting the manipulated objects \cite{hoque2025egodex}. These observations make it possible to recover what the human hands and objects did. They do not, however, specify how a target robot should physically reproduce the interaction.

This distinction is fundamental because transferring dexterous manipulation is not merely a hand-pose mapping problem. Human and robot hands differ in kinematic structure, reachable contact regions, collision geometry, actuation limits, and force capabilities. Moreover, hand--object contacts reconstructed from video are uncertain because of occlusion, depth ambiguity, tracking noise, and geometric interpenetration. Even with a perfectly reconstructed object trajectory, the robot-side interaction remains underdetermined: the same object motion may be produced by many contact topologies, surface contact locations, and force distributions. The human realization is therefore neither fully observed nor necessarily feasible for the target embodiment. We refer to the missing intermediate problem as \emph{physical robot hand--object interaction reconstruction}: recovering one robot-feasible contact--force--motion realization that explains the demonstrated object behavior.

Recent work has tackled this gap at several levels. Geometric and learned retargeting methods transfer human motion into robot references or tracking controllers \cite{park2025transferhandskills,ren2025motiontracks,yin2025geort, liu2025dextrack,li2025maniptrans,zhao2024dexh2r,feng2026regrind,liang2026universal}. Interaction-aware methods additionally preserve contact maps, local hand--object geometry, or interaction topology \cite{lin2025dexflow,wu2026toporetarget,lai2026objretarget, ren2026c2dex}, while physics-aware approaches use simulation, contact dynamics, curricula, or wrench representations to improve feasibility \cite{pan2025spider,zhu2026chord,zhao2025dexmachina,xu2025dexplore, yoo2026softact}. In parallel, end-to-end systems have converted monocular, stereo, or RGB-D human observations into robot trajectories, policies, and training data \cite{zhou2025yoto,lum2025human2sim2robot,chen2026videomanip, hsieh2025dexman,xu2026demobot,niu2026egoaero,han2026video2sim2real, paliwal2026doasido,mu2026deximit}. These results establish that contact and physics matter and that human-observation-to-robot transfer is possible. They do not, however, imply that a concrete robot-side contact--force--motion realization has been explicitly resolved before policy learning.

We present \OURS (\figref{fig:teaser}; Sec.~\ref{sec:method}), a system for reconstructing a physically grounded robot-HOI reference from an \textbf{EGO}centric human demonstration using a new \textbf{C}ontact-aware o\textbf{PT}imization method. The native input consists of a monocular RGB clip and synchronized metric 3D hand skeletons generated from VR devices like Apple Vision Pro or Meta Quest. The perception stage estimates object geometry and SE(3) motion and produces uncertain hand--object contact observations. Rather than treating the reconstructed human contacts as hard constraints, \OURS treats the demonstrated object motion as the task-level invariant and the robot-side contact realization as latent. Its output explicitly instantiates the robot contact topology, object-surface anchors, contact forces, and the corresponding bimanual wrist and joint motion.

Constructing this representation raises a key technical question: how should a robot-side physical realization be selected? \OURS formulates this search as an inverse interaction problem over robot contacts, forces, and motion. Our key insight is that the demonstrated object motion defines the mechanical demands of the task, but does not prescribe a unique grasp or force distribution. We exploit this freedom to search for robot-native interactions, using human hand motion as a spatial prior rather than a contact prescription. More importantly, contact alone is not enough. The robot must also apply the forces needed to reproduce the object motion. We therefore couple contact selection, hand-pose refinement, and force allocation (Sec.~\ref{sec:method_optimization}) to reproduce a robot-feasible interaction, rather than merely a human-like hand configuration. In summary, the contributions of this work are as follows:
\begin{itemize}
    \item \textbf{Physical robot-HOI reconstruction.}
    We formulate human-to-robot transfer as recovering an explicit robot contact--force--motion realization of the demonstrated object motion, rather than directly preserving human hand poses or contacts.

    \item \textbf{Embodiment-aware interaction search.}
    We construct bimanual robot references through contact sampling, hand-pose refinement, and force allocation. Candidate interactions are evaluated by whether they can reproduce the wrench required by the demonstrated object motion while satisfying kinematic, collision, and actuator constraints.

    \item \textbf{Cross-dataset and real-world validation.}
    We evaluate robot-HOI reconstruction across five manipulation task families using noisy visual reconstructions from EgoDex and clean motion-capture data from TACO, and demonstrate bimanual trajectory replay on a real robot system.
\end{itemize}
\section{Related Work}
\label{sec:related_work}




\subsection{Contact and Physics Grounding for Dexterous Transfer}

Interaction-aware methods move beyond hand-pose alignment by preserving object-relative contact structure. DexFlow optimizes temporally consistent robot motion under contact-map constraints \cite{lin2025dexflow}. TopoRetarget preserves a sparse hand--object interaction graph through Laplacian deformation \cite{wu2026toporetarget}, while ObjRetarget maintains multifinger geometric invariants together with arm constraints \cite{lai2026objretarget}. C2Dex aggregates noisy frame-wise estimates into stable object-side contacts and uses the resulting representation for both human-HOI reconstruction and robot retargeting \cite{ren2026c2dex}. Analyses of retargeting objectives likewise show that the selected geometric invariant strongly affects downstream dexterous behavior \cite{xin2025retargetobjectives}. These works establish that object-relative interaction is a more useful transfer signal than hand pose alone.

Physics-aware methods further account for the dynamic consequences of contact. SPIDER uses physics-based sampling and virtual contacts to turn kinematic demonstrations into dynamically feasible robot trajectories \cite{pan2025spider}. CHORD represents the functional capability of an interaction in object-centric wrench space and uses it to guide reinforcement-learning exploration \cite{zhu2026chord}. SoftAct transfers measured human contact and force information to a soft robotic hand \cite{yoo2026softact}, whereas DexSynRefine synthesizes interaction motion and adapts it through residual learning and contact dynamics \cite{lee2026dexsynrefine}. RoboPaint similarly uses geometry- and force-guided tactile retargeting, but relies on a dedicated multimodal capture setup with measured tactile information \cite{fan2026robopaint}. Together, these methods demonstrate that contact geometry alone is insufficient. To solve that, \OURS treats object motion as the goal and hand poses as a reference for finding physically feasible robot contacts, as detailed in Sec.~\ref{sec:method_optimization}. 

\subsection{Robot Skills and Demonstration Data from Human Observations}

A growing body of work converts human observations directly into robot skills. From RGB or stereo video, YOTO extracts keyframes and synthesizes bimanual demonstrations \cite{zhou2025yoto}; VideoManip reconstructs hand--object trajectories, refines geometric contact, and trains a task policy \cite{chen2026videomanip}; DexMan learns bimanual skills from reconstructed third-person videos \cite{hsieh2025dexman}; and Do as I Do uses dynamics-aware optimization to produce robot-complete trajectories from everyday videos \cite{paliwal2026doasido}. DexImit generates physically feasible bimanual episodes from monocular videos \cite{mu2026deximit}, while Dexterous Point Policy learns point-based control without explicitly reconstructing complete meshes or contact forces \cite{kim2026dexterouspointpolicy}. HOWTransfer recovers contact-localized trajectories for parallel-jaw manipulation \cite{shi2026howtransfer}, and Human2Any emphasizes compositional object--object interaction and robot-feasibility reasoning \cite{cheng2026human2any}.

RGB-D systems reduce metric-depth ambiguity and support more direct scene construction. Human2Sim2Robot uses an RGB-D demonstration to define object-motion rewards and exploration priors for sim-to-real reinforcement learning \cite{lum2025human2sim2robot}. DemoBot extracts and refines bimanual motion priors from third-person RGB-D video \cite{xu2026demobot}; EgoAERO reconstructs an asset-free interaction from a single egocentric RGB-D video and trains a residual policy \cite{niu2026egoaero}; and Video2Sim2Real builds a digital twin and combines keyframe optimization, imitation learning, motion planning, and residual reinforcement learning \cite{han2026video2sim2real}. These systems show that raw human video can support complete observation-to-robot pipelines. Our difference is an explicit robot-side physical representation inserted between perception and policy learning, as illustrated in \figref{fig:teaser} and detailed in Secs.~\ref{sec:method_optimization}--\ref{sec:method_rl}.

\section{Method}
\label{sec:method}

\begin{figure*}[t]
 \centering
 \includegraphics[width=\textwidth,height=0.42\textheight,keepaspectratio]{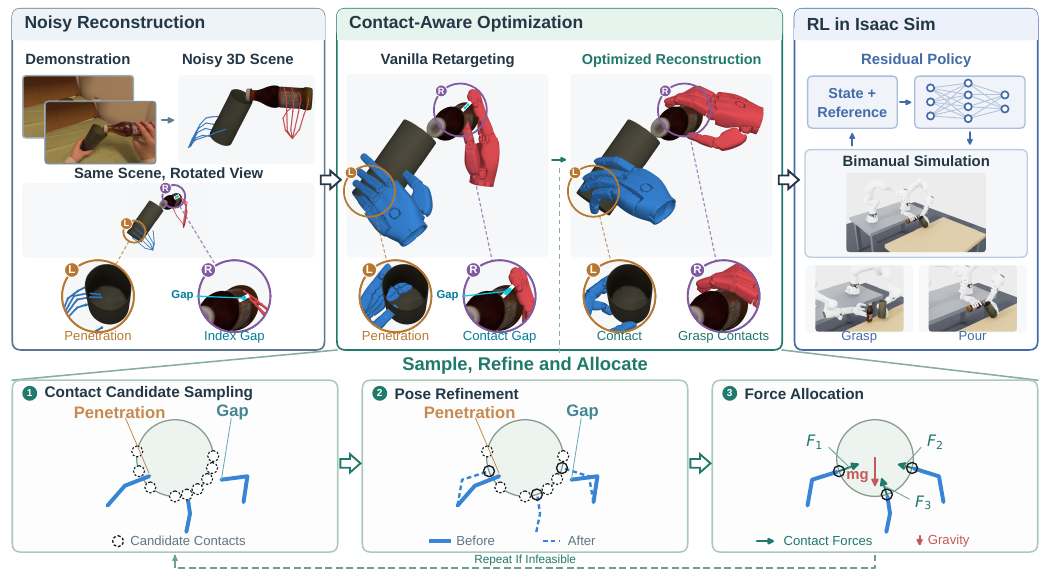}
 \caption{Overview of \OURS. \emph{Left:} Noisy Reconstruction recovers object motion and a hand reference from a demonstration; rotated views expose penetrations and contact gaps. \emph{Center:} Contact-Aware Optimization searches contacts and alternates pose refinement with force allocation while preserving the object-motion goal. \emph{Right:} Reference-conditioned reinforcement learning (RL) uses the resulting robot reference in bimanual simulation. Insets L use a cup-mouth view, and insets R expose finger--bottle geometry. Contact comparisons show the same frame; force sketches are schematic, and Isaac Sim snapshots illustrate the execution setting.\label{fig:pipeline}}
\end{figure*}

\OURS separates \emph{what the objects should do} from \emph{how the robot hands should make it happen} (\figref{fig:pipeline}). Off-the-shelf reconstruction produces noisy scenes: apparent hand--object contact can become penetration or separation, so copying the reconstructed interaction is not a reliable route to policy learning. We therefore treat \hyperref[sec:method_setup]{object trajectories as fixed goals and hand motion as a reference}, not a contact prescription. Guided by that reference, we \hyperref[sec:method_optimization]{search for robot poses and contacts} that can physically support the goal motion. This design resolves the interaction before policy learning while leaving \hyperref[sec:method_rl]{closed-loop reinforcement learning (RL)} to realize it under simulated dynamics.

\subsection{Noisy Reconstruction}
\label{sec:method_setup}

The goal of the reconstruction stage is to extract object trajectories from the provided RGB video and reference hand poses; however, this reconstruction stage is inherently noisy and not physically grounded, often resulting in unwanted penetrations or gaps between the fingers and objects, as shown in Fig.~\ref{fig:pipeline}. We use FoundationPose~\cite{wen2024foundationpose} to obtain the trajectory of the object, which requires three inputs: (1) a mask of the object, (2) a mesh of the object, and (3) an RGBD video. We obtain the object mask using SAM~3~\cite{carion2025sam3segmentconcepts}, the object mesh using SAM~3D~Objects~\cite{sam3dteam2025sam3d3dfyimages}, and the RGBD video using ViGeo~\cite{yu2026vigeo}. ViGeo takes the original RGB image as input and generates the corresponding RGBD video. However, the mesh returned by SAM 3D does not contain dimensional information, and the depth information from ViGeo is noisy. Therefore, we scale the RGBD video and the object mesh according to the reference hand poses. The resulting object trajectories, together with the reference hand poses, are then used in the next stage to retarget the motion to robotic hands while maintaining physical grounding through contact-aware optimization.

\subsection{Contact-Aware Optimization}
\label{sec:method_optimization}

\makeatletter
\newcommand{\optimizationanchor}[1]{%
  \phantomsection
  \begingroup
  \protected@edef\@currentlabel{\p@subsection\thesubsection}%
  \label{#1}%
  \endgroup
}
\makeatother


The vanilla retargeting~\cite{sharpa_manus_sdk} step (\figref{fig:pipeline}, center-left) retargets the human hand pose trajectory to the robot hand trajectory. This linkage-optimization-based retargeting algorithm does not address the lack of physical grounding from the noisy reconstruction stage, and therefore maintains the unwanted penetrations and gaps. However, it provides a good starting point for applying our contact-aware optimization pipeline, which iteratively performs contact candidate sampling, pose refinement, and force allocation, as elaborated below. The contact-aware optimization pipeline follows the idea of rejection sampling, where sampled contacts must survive the subsequent two stages; otherwise, the sampling process continues with new contact candidates. 

\textbf{Contact Candidate Sampling.}\optimizationanchor{sec:method_search}
In the contact candidate sampling stage, we first randomly sample points on the object surface and near the original endpoints of the robot fingers. We denote these sampled points as $p \in \mathbb{R}^{30}$, representing the 3-DoF end positions of ten fingers. Note that these sampled points are only potential candidates and are not guaranteed to be valid contact locations. As shown in~\figref{fig:pipeline}, their validity can only be determined in the later pose refinement and force allocation stages.

The iterative contact-aware optimization process runs in parallel across all frames, and the sampled set of candidates is not guaranteed to be the same across different frames. If a contact candidate is validated in a later frame, we propagate it to the other frames as well.

\optimizationanchor{sec:method_acquisition}


\textbf{Pose Refinement.}\optimizationanchor{sec:method_pose}
Given a contact candidate $p \in \mathbb{R}^{30}$ sampled from the sampling stage, we define $x\in\mathbb{R}^{56}$ as the state of the hands, consisting of 22 DoF for the left hand, 22 DoF for the right hand, 6 DoF for the left wrist pose, and 6 DoF for the right wrist pose. Note that, at any video frame $t$, $x_t$ can either be the reference hand poses $x_r$ returned by the vanilla retargeting algorithm, which may not be physically grounded, or a valid solution obtained propagated back from another frame $x_m$, where $m \neq t$. The goal of the pose refinement stage is to compute hand joint movements $\delta x$ that make the contact candidate viable. If such a movement is found, we proceed to the force allocation stage; otherwise, we move on to the next sampled contact candidate.

The optimization problem we solve is defined in Equation~\ref{eq:pose_update}. Here, $f_{\text{FK}}(x)$ denotes the forward kinematics function that maps $x$ to the finger endpoints in Euclidean space. Note that $x$ contains not only the joint values but also the poses of the two wrists, which require additional handling; we omit these details for simplicity. The function $d(x)$ computes the penetration distance of the fingers into the object surface.

The optimization objective consists of three terms: (1) after applying the finger displacement $\delta x$, the resulting finger endpoints $f_{\text{FK}}(x + \delta x)$ should be close to the contact candidate $p$, encouraging $\|f_{\text{FK}}(x + \delta x) - p\|^{2}$ to be small; (2) the updated finger state $x + \delta x$ should remain close to the reference state $x_r$, encouraging $\|x + \delta x - x_r\|^{2}$ to be small; and (3) the displacement $\delta x$ should remain small in general, enforced by the regularization term $\|\delta x\|^{2}$. Note that when the current hand state satisfies $x = x_r$ (i.e., $x$ is not propagated from another frame), the second and third terms are equivalent. The constraints ensure that the hand joints remain within their respective joint limits and that the penetration distance remains below a predefined threshold $\epsilon$.


\begin{equation}
\begin{aligned}
 \min_{\delta x}\quad &
 \alpha \|f_{\text{FK}}(x + \delta x) - p\|^{2} 
 +\beta \|x + \delta x - x_r\|^{2}
 +\eta \|\delta x\|^{2}\\
 \mathrm{s.t.}\quad &
 x_{\min} \leq x+\delta x \leq x_{\max},\\
 &d(x+\delta x) \geq-\epsilon
\end{aligned}
\label{eq:pose_update}
\end{equation}


After pose refinement, we smooth consecutive hand states using Equation~\ref{eq:temporal_correction}, subject to the same joint-limit and penetration constraints (omitted for brevity). Here, $\alpha$, $\beta$, and $\eta$ are hyperparameters that control the relative weights of the different optimization terms, and $b$
specifies per-coordinate correction limits, restricting smoothing
to a local neighborhood of the refined reference.



\begin{equation}
\begin{aligned}
\min_{\delta x_t}\quad
& \sum_{t=1}^{N-1}
\left\|
(x_t+\delta x_t)-(x_{t-1}+\delta x_{t-1})
\right\|_2^2 \\
\text{s.t.}\quad
& -b \leq \delta x_t \leq b
\end{aligned}
\label{eq:temporal_correction}
\end{equation}

\textbf{Force Allocation.}\optimizationanchor{sec:method_forces}\optimizationanchor{sec:method_wrench}
After pose refinement, force allocation tests whether the candidate contacts can support the object's reference motion (\figref{fig:pipeline}, center-bottom). To determine this, we enforce the sum of all contact wrenches to match the object wrench required to realize the target trajectory.  The optimization problem we solve is defined in Equation~\ref{eq:force_qp}. 

\optimizationanchor{sec:method_support}Here, the matrix $\mathbf{G}\in\mathbb{R}^{6\times K}$ is the grasp matrix constructed from both finger--object and environment--object contact points, where $K$ denotes the total number of contact-force basis directions. For environmental contacts, we infer a stationary support disk from the bottom of each object's collision geometry at its initial pose. We first check whether support alone can sustain the reference motion, then allocate the required wrench among the available finger and support contacts. Each column of $\mathbf{G}$ contains a force direction within the corresponding friction cone and its induced moment about the object's center of mass. The vector $\boldsymbol{\lambda}\in\mathbb{R}^{K}$ collects the corresponding coefficients, with $\boldsymbol{\lambda}\geq\mathbf{0}$ enforcing unilateral contact and friction constraints. The resultant contact wrench is $\mathbf{G}\boldsymbol{\lambda}$. The target $w_t^{\mathrm{req}}\in\mathbb{R}^{6}$ follows from the Newton--Euler equations using the reference object motion, mass, inertia, and gravity.

For each object, we solve the following wrench-matching objective subject to the maximum allowable contact forces. Here, $\rho>0$ controls the regularization strength of the force coefficients:
\begin{equation}
\begin{aligned}
\min_{\boldsymbol{\lambda}}\quad
& \|\mathbf{G}\boldsymbol{\lambda}
-\hyperref[sec:method_wrench]{w_t^{\mathrm{req}}}\|_2^2
+\rho\|\boldsymbol{\lambda}\|_2^2 \\
\text{s.t.}\quad
& \boldsymbol{\lambda}\geq\mathbf{0},\quad
\|\mathbf{f}_i(\boldsymbol{\lambda})\|_2\leq f_i^{\max},\quad\forall i.
\end{aligned}
\label{eq:force_qp}
\end{equation}

Here, $\mathbf{f}_i(\boldsymbol{\lambda})$ is the force at contact $i$, obtained by summing its force basis directions weighted by the corresponding coefficients in $\boldsymbol{\lambda}$. The bound $f_i^{\max}$ limits how much force each contact can exert. The first term penalizes wrench mismatch, while the second regularizes the force coefficients, discouraging unnecessarily large forces, including opposing forces whose contributions cancel.

\optimizationanchor{eq:shared_effort}We accept a candidate only if its wrench residual is below the prescribed threshold and the combined robot loads from all objects satisfy the shared actuator limits. Otherwise, we return to contact search.

\subsection{Residual Reinforcement Learning}
\label{sec:method_rl}

Execution errors can alter contacts and prevent the forces found by contact-aware optimization from being realized. We therefore train a single residual policy to adjust both hands around the optimized reference and track both object trajectories (\figref{fig:pipeline}, right).

\textbf{Residual Policy.}
The reference provides a desired hand motion, and the policy learns small wrist and finger corrections to maintain useful contacts and track the object. Object poses and reference targets indicate tracking progress. Recent wrist and finger positions, the previous action, and sequence time describe the robot's motion and reference phase. A Gaussian policy maps these observations to $u=[u_{p}^{\top},u_{R}^{\top},u_{q}^{\top}]^{\top}\in\mathbb{R}^{56}$ stacks 6 wrist-translation, 6 wrist-rotation, and 44 joint corrections for both hands. For simplicity, action scaling and clipping are absorbed into $u$. Applied to the current reference wrist rotation matrix $R_{\mathrm{ref}}$, position $p_{\mathrm{ref}}$, and finger-joint configuration $q_{\mathrm{ref}}$, they yield
\begin{equation}
\begin{aligned}
 p_{\mathrm{cmd}}&=p_{\mathrm{ref}}
       +u_{p},\\
 R_{\mathrm{cmd}}&=
       \exp([u_{R}]_\times)
       R_{\mathrm{ref}},\\
 q_{\mathrm{cmd}}&=
       \operatorname{clip}_{[q_{\min},q_{\max}]}
       (q_{\mathrm{ref}}+u_{q}).
\end{aligned}
\label{eq:rl_residual_targets}
\end{equation}
Here, $[\cdot]_\times$ denotes the cross-product matrix; clipping enforces the lower and upper finger-joint limits $q_{\min}$ and $q_{\max}$. Offsets are applied independently to each reference frame without accumulation. Wrist impedance and finger position drives track these pose targets, so the residual actions influence object motion through contact forces.

\textbf{Tracking and Contact Rewards.}
We design the reward around a simple principle: the object should follow the reference motion, and the contacts should provide the physical wrench needed to make that motion happen. For object tracking, we compare corresponding keypoints on the actual and reference objects, using distance thresholds that adapt to object size and pose uncertainty, while giving less weight to unreliable frames. We also encourage the object to move away from its initial pose and to match the reference velocity, with a weaker wrist-tracking term used only as additional guidance. When the object pose is poorly tracked, velocity rewards are reduced so that the policy first focuses on recovering the pose. For contact, the policy is rewarded for making contact when the reference indicates it should, and for producing contact forces and moments that match the wrench implied by the reference acceleration and gravity. Bounded feedback from object pose and velocity errors adjusts the target wrench to aid recovery. Finally, we penalize large or jerky actions, actions beyond the clipping range, excessive contact forces, and tracking failures.

\textbf{Policy Training.}
We train with proximal policy optimization (PPO)~\cite{schulman2017ppo} and a critic observing noiseless simulation state, fingertip positions, and joint velocities. The residual is initialized to zero. Episodes start at the first frame or adaptively selected later frames, with randomization of friction, masses, controller gains, joint offsets, and observation delays. A curriculum increases gravity from zero to $9.81$\,m/s$^2$ according to the recent success rate of complete bimanual episodes. Success requires both objects to complete their goal trajectories within the translation, rotation, and keypoint-error bounds; final evaluation uses the same criterion under normal gravity. After that, we replayed the trajectory generated from the residual policy directly on the real robot.

\section{Experiments}
\label{sec:experiments}
\providecommand{\resultNA}{\textemdash}

We evaluate \OURS across five manipulation tasks on noisy EgoDex reconstructions and cleaner TACO motion-capture inputs. Our experiments have three objectives. First, we evaluate how effectively \OURS reproduces demonstrated object motions compared with existing methods. Second, we assess the contribution of force allocation to execution after RL by comparing reference variants under identical training configurations and budgets. Third, we evaluate robustness to hand-position noise at both the reference and execution stages, measuring penetration, contact/wrench feasibility, and execution failures after RL.

\begin{table}[!t]
\centering
\caption{Success rates (\%, $\uparrow$) on the 52 noisy EgoDex
reconstructions.}
\label{tab:egodex}
\footnotesize
\setlength{\tabcolsep}{2pt}
\renewcommand{\arraystretch}{1.12}
\begin{tabular*}{\columnwidth}{@{\extracolsep{\fill}}lcccccc@{}}
\toprule
Method & \makecell{Pouring}
       & \makecell{Lint\\rolling}
       & \makecell{Sweep\\dustpan}
       & \makecell{Scoop\\ice}
       & \makecell{Pick \&\\place}
       & \makecell{Overall} \\
\midrule
CHORD                    & \textbf{70.6} & 37.5 & 45.5 & \textbf{85.7} & 33.3 & 55.8 \\
SPIDER   & 11.8 &  0.0 & 18.2 &  14.3 & 33.3 & 15.4 \\
Do As I Do & 64.7 & 12.5 & 18.2 & 57.1 & 66.7 & 46.2 \\
Ours       & \textbf{70.6} & \textbf{62.5} & \textbf{54.5} & 71.4 & \textbf{77.8} & \textbf{67.3} \\
\bottomrule
\end{tabular*}
\end{table}

\begin{figure*}[!t]
\centering
\includegraphics[width=\textwidth]{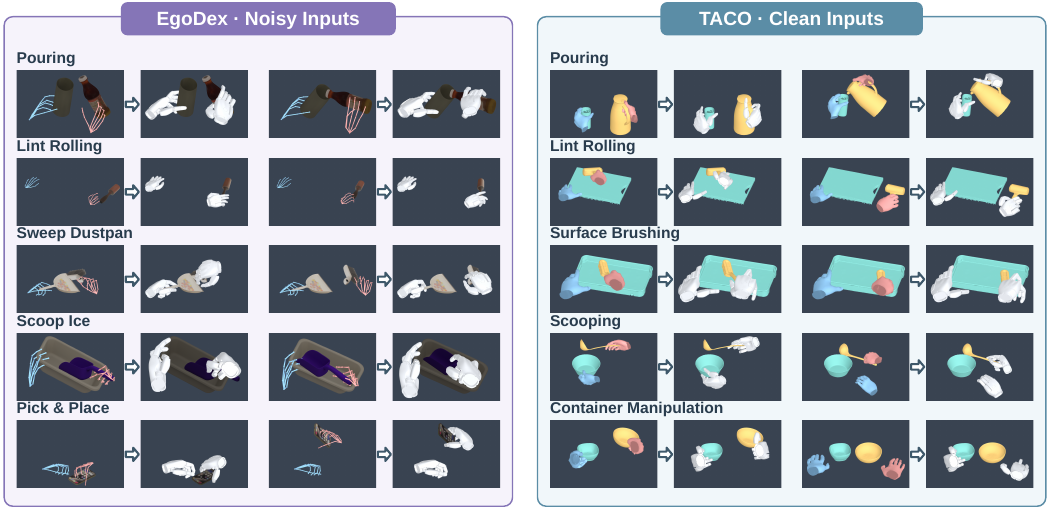}
\caption{\textbf{Input-to-output comparisons on EgoDex and TACO.} The left and right groups show EgoDex and TACO, respectively. Within each pair, the input (left) consists of noisy object reconstructions and reference human hand skeletons for EgoDex, or motion-capture hand--object data for TACO; the output (right) shows the interaction reconstructed by \OURS with white Sharpa hands. From top to bottom, EgoDex rows show pouring, lint rolling, sweeping with a dustpan, scooping, and pick and place; TACO rows show pouring, lint rolling, surface brushing, scooping, and container manipulation.}
\label{fig:input_output_comparison}
\end{figure*}

\begin{table}[!t]
\centering
\caption{Success rates (\%, $\uparrow$) on the 25 clean TACO
reconstructions, with five sequences per task group. The last three groups are analogous tool--object
tasks.}
\label{tab:taco}
\footnotesize
\setlength{\tabcolsep}{2pt}
\renewcommand{\arraystretch}{1.12}
\begin{tabular*}{\columnwidth}{@{\extracolsep{\fill}}lcccccc@{}}
\toprule
Method & \makecell{Pouring}
       & \makecell{Lint\\rolling}
       & \makecell{Surface\\Brushing}
       & \makecell{Scooping}
       & \makecell{Container\\Manipulation}
       & \makecell{Overall} \\
\midrule
CHORD & 60.0 & 40.0 & 80.0 & 0.0 & \textbf{80.0} & 52.0 \\
SPIDER     & 80.0 & 80.0 & 40.0 & 80.0 & 60.0 & 68.0 \\
Do As I Do &  0.0 & 20.0 & 20.0 &  0.0 & 20.0 & 12.0 \\
Ours       & \textbf{100.0} & \textbf{100.0} & \textbf{100.0} & \textbf{100.0} & 60.0 & \textbf{92.0} \\
\bottomrule
\end{tabular*}
\end{table}

\begin{figure*}[!t]
\centering
\begin{minipage}[b]{0.32\textwidth}
\centering
\includegraphics[width=\linewidth]{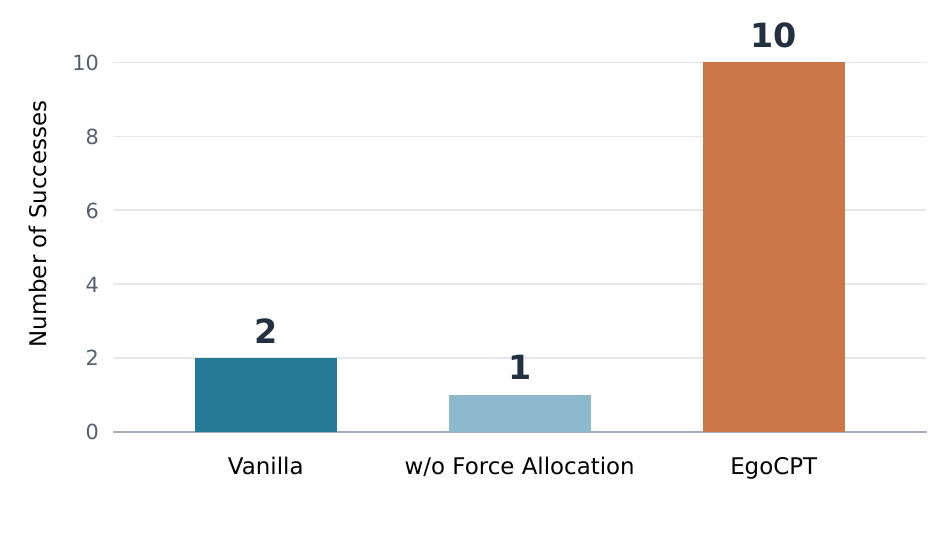}
\par\small (a) The Importance of Force Allocation
\end{minipage}\hfill
\begin{minipage}[b]{0.32\textwidth}
\centering
\includegraphics[width=\linewidth]{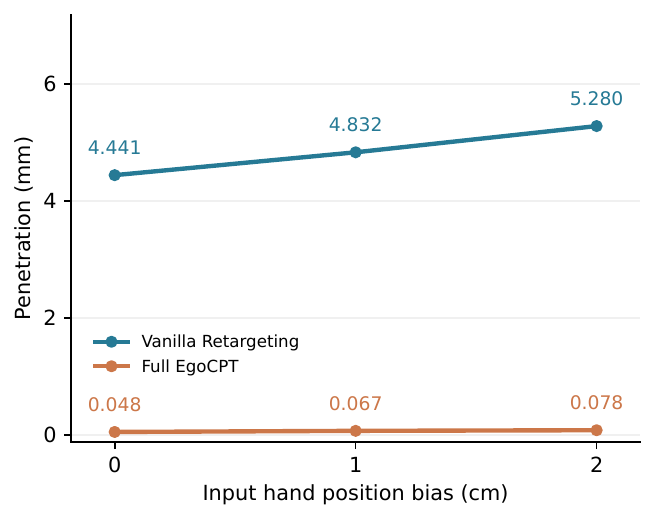}
\par\small (b) Penetration under position noise
\end{minipage}\hfill
\begin{minipage}[b]{0.32\textwidth}
\centering
\includegraphics[width=\linewidth]{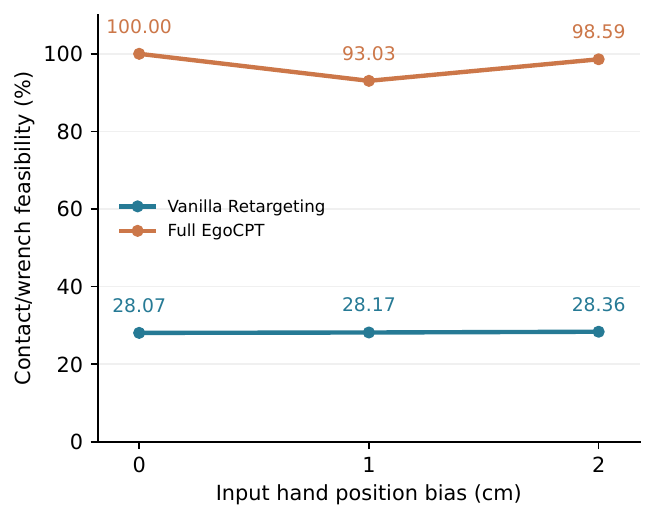}
\par\small (c) Contact/wrench feasibility
\end{minipage}
\caption{\textbf{The Importance of Force Allocation and robustness to hand-position noise.} (a) Successful executions after RL on 10 EgoDex sequences (2 per task group), with identical training configurations and budgets across all three reference variants. (b,c) Reference penetration and frame-level contact/wrench feasibility on 15 TACO sequences (3 per task group), with a constant translation of 0, 1, or 2\,cm applied throughout each input hand trajectory. Feasibility is pooled over all frames in all evaluation videos; accepted frames must solve successfully and satisfy the contact-validity and wrench-residual thresholds.}
\label{fig:ablation_robustness}
\end{figure*}

\subsection{Experimental Setup}
\label{sec:exp_benchmarks}
\label{sec:exp_metrics}

The noisy benchmark contains fifty-two EgoDex~\cite{hoque2025egodex} sequences: seventeen pouring, eight lint rolling, eleven sweeping, seven scooping, and nine pick-and-place sequences. We exclude upstream reconstructions with incorrect object pose, although retained inputs may still contain contact misalignment and dynamic inconsistencies. Each task is defined by the demonstrated motion of its manipulated rigid objects.

The motion-capture benchmark contains twenty-five TACO~\cite{liu2024taco} sequences, with five examples each of pouring, lint rolling, surface brushing, scooping, and container manipulation. We use the provided object meshes and hand--object data captured through the same interface described in Sec.~\ref{sec:method_setup}. Together, the two benchmarks evaluate transfer from visual reconstructions and motion-capture demonstrations. We use the Sharpa Hand in all experiments.

We compare with CHORD~\cite{zhu2026chord}, which uses contact-wrench guidance for policy learning; SPIDER~\cite{pan2025spider}, which uses physics-based sampling for retargeting; and Do As I Do~\cite{paliwal2026doasido}, which converts reconstructed interactions into robot actions. Baselines receive the human-hand and object reconstructions at the \emph{Noisy 3D Scene} stage (\figref{fig:pipeline}, left), retaining required input adaptation and native preprocessing.

We report success as the percentage of accepted sequences in each evaluation set, using numerical thresholds of $0.10$\,m for position and $0.50$\,rad for rotation across all four methods. Tables~\ref{tab:egodex} and~\ref{tab:taco} report the results for EgoDex and TACO.

\subsection{Comparison and Analysis}
\label{sec:exp_results}

On noisy EgoDex reconstructions, \OURS achieves 67.3\% success, improving on CHORD, the strongest baseline, by 11.5 percentage points (Table~\ref{tab:egodex}). Inspection of failed sequences suggests that noisy contacts corrupt CHORD's reference wrench guidance, consistent with its reported sensitivity to contact noise~\cite[Appendix~A.1]{zhu2026chord}. SPIDER's virtual contact guidance forces steer optimization toward reference hand--object contacts~\cite{pan2025spider}; when these contacts are reconstructed incorrectly, the forces can guide the search toward incorrect interactions and cause failure. \OURS treats human contacts as uncertain observations and searches for robot contacts that can support the demonstrated object motion. The advantage spans several task families, although CHORD remains stronger on scooping.

\label{sec:exp_clean}

On cleaner TACO inputs, \OURS achieves 92.0\% success, exceeding SPIDER, the strongest baseline, by 24.0 percentage points (Table~\ref{tab:taco}). When reconstructed contacts are reliable, SPIDER's virtual guidance forces instead help establish the intended interaction, consistent with its stronger performance on TACO. Failure inspection reveals a different issue for Do As I Do: despite its noise-handling mechanisms, the evaluated implementation identifies a grasp whenever the hand--object distance is below 10\,cm. This rule can mistake proximity for grasping and bias the recovered interaction away from a clean reference. Accurate geometry alone therefore does not guarantee correct contact assignment. \OURS evaluates candidate contacts against the forces and motion required by the task. Its two remaining failures occur in container manipulation, where CHORD performs better.

\label{sec:exp_cases}

To examine the recovered interactions, \figref{fig:input_output_comparison} compares inputs and outputs across five task groups on EgoDex (left group) and TACO (right group), illustrating how \OURS adapts human hand placements to robot contacts. Trajectory traces distinguish two failure patterns. On EgoDex lint rolling (\figref{fig:input_output_comparison}, left group, second row), CHORD terminates after four reference frames, with object position error reaching 22.3\,cm. On TACO pouring (\figref{fig:input_output_comparison}, right group, first row), Do As I Do completes the sequence, but the tool's mean errors of 11.2\,cm and 0.570\,rad exceed both acceptance thresholds, while the receiving object's mean errors remain within the bounds. \OURS produces an accepted reconstruction in both cases. These cases distinguish early tracking breakdown from inaccurate execution despite completion.

Environmental support also affects contact feasibility. On a 223-frame TACO sequence, revised support inference and contact-search settings reduce infeasible frame--object pairs from forty-five to zero with the input meshes and object-motion goals fixed. This case illustrates how support determines which loads the robot contacts must supply (Sec.~\ref{sec:method_forces}). Since both support handling and contact search changed, the result does not isolate their individual contributions.

\subsection{The Importance of Force Allocation}
We test whether force allocation improves execution after RL by comparing vanilla retargeting, our reference without force allocation, and full EgoCPT. All variants use the same 10 EgoDex sequences (two from each of five task groups), RL training configurations, and budgets.

As shown in \figref{fig:ablation_robustness}(a), execution success after RL is 10/10 (100\%) for full EgoCPT, 2/10 (20\%) for vanilla retargeting, and 1/10 (10\%) without force allocation. On these sequences, contact and pose refinement without force allocation does not improve execution over vanilla retargeting. The improvement with full EgoCPT supports the role of force allocation in constructing effective references for policy learning.

\subsection{Robustness to Hand-Position Noise}
We evaluate reference quality and execution after RL on 15 TACO sequences (three from each of five task groups), applying a fixed 0, 1, or 2 cm translation to each entire input hand trajectory. We compare vanilla retargeting with full EgoCPT using reference penetration and frame-level contact/wrench feasibility (\figref{fig:ablation_robustness}(b,c)). Feasibility is the fraction of all frames across all evaluation videos for which optimization succeeds and both the actual-contact validity check and the fixed wrench-residual threshold are satisfied. Failed frames remain in the denominator.

Full EgoCPT maintains low penetration and high contact/wrench feasibility across the tested noise levels. At 0, 1, and 2 cm, its penetration values are 0.048, 0.067, and 0.078 mm, respectively, compared with 4.441, 4.832, and 5.280 mm for vanilla retargeting. The corresponding feasibility rates are 100.00\%, 93.03\%, and 98.59\% for full EgoCPT, versus 28.07\%, 28.17\%, and 28.36\% for vanilla retargeting. These results show that contact-aware optimization preserves reference quality under the tested input perturbations.

Execution after RL remains sensitive to increasing noise. At 0, 1, and 2 cm, full EgoCPT succeeds on 15/15 (100\%), 13/15 (86.7\%), and 11/15 (73.3\%) sequences, respectively. Thus, high aggregate frame-level feasibility alone does not ensure successful execution of a full sequence.

\section{Conclusion \& Limitation}
We presented \OURS, a system for learning dexterous manipulation from a single egocentric RGB video and reference hand poses using a novel contact-aware optimization framework. It iterates between contact candidate sampling, pose refinement, and force allocation to reconstruct robot-feasible interactions that preserve the demonstrated object motion. Residual reinforcement learning uses these references for closed-loop execution. Experiments on noisy EgoDex reconstructions and cleaner TACO reconstructions show higher overall success than baselines, and we also deploy this learned trajectory onto a bimanual setup with Sharpa hands.

The remaining failures limit what these experiments establish. Contact search is local and budgeted, so rejection does not prove that no feasible robot interaction exists. The current pipeline assumes rigid objects and relies on reconstructed hand pose trajectories. Future work will improve robustness to upstream reconstruction errors and extend real-world deployment to a wider range of tasks.

\section{Acknowledgment}
OpenAI Codex assisted with code implementation and debugging. All outputs were reviewed by the authors, who take full responsibility for the work.










\bibliographystyle{IEEEtran}
\bibliography{references}

\end{document}